\documentclass[pre,twocolumn,a4paper,superscriptaddress]{revtex4}
\usepackage{amsmath}
\usepackage{amsfonts}
\usepackage{amssymb}
\usepackage{epsfig}
\begin{document}

\title{Cross-situational and supervised learning in the emergence of communication}

\author{Jos\'e F. Fontanari}
\email{fontanari@ifsc.usp.br}
\affiliation{Instituto de F\'{\i}sica de S\~ao Carlos,
  Universidade de S\~ao Paulo,
  Caixa Postal 369, 13560-970 S\~ao Carlos, S\~ao Paulo, Brazil}
   
\author{Angelo Cangelosi}
\email{A.cangelosi@plymouth.ac.uk}
\affiliation{Centre for Robotics and Neural Systems, University of Plymouth, Plymouth PL4 8AA, 
United Kingdom 
}                           
 %\date{sometimes}

\begin{abstract}

Scenarios for the emergence or bootstrap of a lexicon involve the repeated interaction between at least two agents 
who must reach a consensus on how to name $N$ objects using  $H$ words. Here we consider minimal models of 
two types of learning algorithms: cross-situational learning, in which the individuals determine the meaning of a 
word by looking for something in common across all observed uses of that word, and supervised operant conditioning learning, 
in which there is strong feedback between individuals about the intended meaning of the words. Despite the 
stark differences between these learning schemes, we show that they yield the same communication accuracy in the realistic 
limits of large  $N$ and $H$, which coincides with the result of the classical occupancy problem of randomly 
assigning $N$ objects to $H$  words.

\end{abstract}

%\pacs{89.75.Da, 89.75.Fb, 02.50.Ey, 02.50.Le}

\maketitle

\section{Introduction}

How a coherent lexicon can emerge in a group of interacting agents is a major open issue in the language evolution and 
acquisition research area (Hurford, 1989; Nowak \& Krakauer, 1999; Steels, 2002; Kirby, 2002; Smith, Kirby, \& Brighton, 2003). 
In addition, the dynamics in the self-organization of shared lexicons is one of the issues to which computational and mathematical 
modeling can contribute the most, as the emergence of a lexicon from scratch implies some type of self-organization and, possibly, 
threshold phenomenon. This cannot be completely understood without a thorough exploration of the parameter space of the models 
(Baronchelli, Felici, Loreto, Caglioli, \& Steels, 2006).
	
There are two main research avenues to investigate the emergence or bootstrapping of a lexicon. The first approach, 
inspired by the seminal work of Pinker and Bloom (1990) who argued that natural selection is the main design principle 
to explain the emergence and complex structure of language, resorts to evolutionary algorithms to evolve the shared lexicon. 
The key element here is that an improvement on the communication ability of an individual results, in average, in an increase 
of the number of offspring it produces (Hurford, 1989; Nowak \& Krakauer, 1999; Cangelosi, 2001; Fontanari \& Perlovsky, 2007, 2008). 
The second research avenue, which we will follow in this paper, argues for a culturally based view of language evolution and so 
it assumes that the lexicons are acquired and modified solely through learning during the individual's lifetime 
(Steels, 2002; Smith, Kirby, \& Brighton, 2003). 

Of course, if there is a fact about language which is uncontroversial, it is that the lexicon must be learned from the active or 
passive interaction between children and language-proficient adults. The issue of whether this ability to learn the lexicon is 
due to some domain-general learning mechanism, or  is an innate ability, unique to humans, is still on the table (Bates \& Elman, 1996). 
In the problem we address here, there is simply no language-proficient individuals, so it is not so far-fetched to put forward a 
biological rather than a cultural explanation for the emergence of a self-organized lexicon. Nevertheless, in this contribution 
we will use many insights produced by research on language acquisition by children (see, e.g., Gleitman, 1990; Bloom, 2000) to 
study different learning strategies.

From a developmental perspective, there are basically two competing schemes for lexicon acquisition by children 
(Rosenthal \& Zimmerman, 1978). The first scheme, termed cross-situational or observational learning, is based on the 
intuitive idea that one way that a learner can determine the meaning of a word is to find something in common across all 
observed uses of that word (Pinker, 1984; Gleitman, 1990; Siskind, 1996). Hence learning takes place through the statistical 
sampling of the contexts in which a word appears. Since the learner receives no feedback about its inferences, we refer to 
this scheme as unsupervised learning. The second scheme, known generally as operant conditioning, involves the active 
participation of the agents in the learning process, with exchange of non-linguistic cues to provide feedback on the 
hearer inferences. This supervised learning scheme has been applied to the design of a system for communication by 
autonomous robots -- the so-called language game in the Talking Heads experiments (Steels, 2003). Despite the technological 
appeal, the empirical evidence is that most part of the lexicon is acquired by children as a product of unsupervised learning 
(Pinker, 1984; Gleitman, 1990; Bloom, 2000).

Interestingly, from the perspective of evolving or bootstrapping a lexicon, the unsupervised scheme is very attractive too, 
since it eliminates altogether the issue of honest signaling (Dawkins \& Krebs, 1978), as no signaling is involved in the 
learning process, which requires only observation and some elements of intuitive psychology (e.g. Theory of Mind). 

Many different computational implementations and variants of these two schemes for bootstrapping a lexicon have 
been proposed in the literature. For example, Smith (2003a, 2003b), Smith, Smith, Blythe, \& Vogt (2006), 
and De Beule, De Vylder, \& Belpaeme (2006) have addressed the unsupervised learning scheme, 
whereas Steels \& Kaplan (1999), Ke, Minett, Au, Wang (2002), Smith, Kirby, \& Brighton, (2003), 
and  Lenaerts, Jansen,  Tuyls, \& De Vylder (2005), the supervised scheme. However, except for the extensive statistical 
analysis of a variant of the supervised learning algorithm which reduces the problem to that of naming a single object 
(Baronchelli, Felici, Loreto,  Caglioli, \& Steels, 2006), the study of the effects of changing the parameters of those 
models have been usually limited to the display of the time evolution of some measure of the communication accuracy of the 
population. Although at first sight the supervised learning scheme may seem to be clearly superior to the unsupervised 
one (albeit less realistic in the context of language acquisition by children), we are not aware of any thorough 
comparison between the performances of these two learning scenarios. In fact, in this contribution we show that in 
a realistic limit of very large lexicon sizes the supervised and unsupervised learning performances are essentially identical. 

In this paper we study minimal models of the supervised and unsupervised learning schemes which preserve the main 
ingredients of these two classical language acquisition paradigms. For the sake of simplicity, here we interpret the 
lexicon as a mapping between objects and words (or sounds) rather than as a mapping between meanings (conceptual structures) 
and sounds. A more complete scenario would involve first the creation of meanings, i.e., the bootstrapping of an object-meaning 
mapping (Steels, 1996; Fontanari, 2006) and then the emergence of a meaning-sound mapping 
(see, e.g., Smith, 2003a, 2003b; Fontanari \& Perlovsky, 2006).

\section{Model}

Following a common assumption in lexicon bootstrapping models, such as the popular iterated learning model 
(Smith, Kirby, \& Brighton, 2003; Brighton, Smith, \& Kirby, 2005 ), we consider here only two agents who play in turns 
the roles of speaker and hearer. The agents live in a fixed environment composed of $N$ objects and have $H$ words available 
to name these objects. As we are interested in the limit where  $N$ and $H$  are very large with the ratio  $\alpha \equiv H/N$ 
finite we do not need to  account for the possibility of creation of new words as in some variants of the supervised learning scheme 
(Baronchelli, Felici, Loreto, Caglioli, \& Steels, 2006).

We assume that each agent is characterized by a  $N \times H$ verbalization matrix  $P$  the entries of 
which $p_{nh} \in \left [ 0,1 \right ]$, with $p_{nh} \in \left [ 0,1 \right ]$  for all values of $n=1,\ldots,N$, 
being interpreted as the probability that object $n$  is associated with word $h$. This assumption rules out the existence 
of objects without names, but it allows for words which are never used to name objects. To describe the communicative behavior of 
the agents through the verbalization matrix (i.e., the associations between objects and words for use both in production and 
interpretation) we need to specify how the speaker chooses a word for any given object as well as how the hearer infers the 
object the speaker intended to name by that word.

To name an object, say object $n$, the speaker simply chooses the word $h^*$  which is associated to the largest entry of row $n$ of the matrix $P$, 
i.e., $h^* = \max_{h} \left \{ p_{nh}, h = 1, \ldots, H \right \}$. In addition, to guess which object the speaker named by word $h$  
the hearer selects the object that corresponds to the largest of the $N$ entries  $p_{nh}$, $n=1, \ldots, N$.
In other words, the hearer chooses the object that it itself would be most likely to associate with word $h$  (Smith, 2003a, 2003b). 
This amounts to assuming that the agents are endowed with a `Theory of Mind' (ToM), i.e., that the hearer is somehow able to understand 
that the speaker thinks similar to itself and hence would behave likewise when facing the same situation (Donald, 1991). 
We note that the original inference scheme, termed ``obverter'' (Oliphant \& Batali, 1997), assumed that the hearer has access to 
the verbalization matrix of the speaker (through mind reading, as the critics were ready to point out). Here we follow the more 
reasonable scheme, dubbed ``introspective obverter'' (Smith, 2003a), which requires endowing the agents with a Theory of Mind 
rather than with telepathic abilities.

Effective communication takes place when the two agents reach a consensus on which word must be assigned to each object. 
To achieve this, we must provide a prescription to modify their initially random verbalization matrices. Here we will consider 
two learning procedures that differ basically on whether the agents receive feedback (supervised learning) or not (unsupervised learning) 
about the success of a communication episode. But before doing this we need to set up the language game scenario where the agents interact. 

From the list of  $N$ objects, the agent who plays the speaker role chooses randomly $C$ objects without replacement. This set of  $C$ objects 
forms the context. Then the speaker chooses randomly one object in the context and produces the word associated to that object, 
according to the procedure sketched before. The hearer has access to that word as well as to the  $C$ objects that comprise the context. 
Its task is to guess which object in the context is named by that word. This is then an ambiguous language acquisition scenario in 
which there are multiple object candidates for any word. Once the verbalization matrices are updated the two agents interchange the 
roles of speaker and hearer and a new context is generated following the same procedure. 

To control the convergence properties of the learning algorithms described next we assume that the entries $p_{nh}$  are discrete variables 
that can take on the  values $0,1/M,2/M,\ldots,1-1/M,1$. In our simulations we choose $M=10^4$. The reciprocal of $M$  can be interpreted as 
the algorithm learning rate. In addition, as there are two agents who alternate in the roles of speaker and hearer, henceforth we will 
add the superscripts I or J to the verbalization matrix in order to identify the agent it corresponds to. At the beginning of the 
language game each agent has a different, randomly generated verbalization matrix. More pointedly, to generate the row $n$  of $P^I$
we distribute with equal probability  $M$ balls among $H$  slots and set the value of entry  $p_{nh}^I$ as the ratio between the number of balls 
in slot $h$  and the total number of balls $M$. An analogous procedure is used to set the initial value of $P^J$.

\subsection{Unsupervised learning}

In this scheme, the list of objects in the context  $n_1, \ldots, n_c$ and the accompanying word $h^*$  is the only information fed to 
the learning algorithm. Hence, in the unsupervised scheme, only the hearer's verbalization matrix is updated. Of course, since 
the agents change roles at each learning episode, the verbalization matrices of both agents are updated during the learning stage. 
For concreteness, let us assume that agent $I$  is the speaker and so agent $J$  is the hearer in a particular learning episode. 
As pointed out before, the idea here is to model the cross-situational learning scenario (Siskind, 1996) in which the agents 
infer the meaning of a given word by monitoring its occurrence in a variety of contexts. Accordingly, the learning procedure 
increases the entries  $p_{n_1 h^*}^J, \ldots, p_{n_c h^*}^J$ by the amount $1/M$. In addition, for each object 
in the context, say $n_1$, a word, say $h$, is chosen randomly and the entry $p_{n_1 h}^J$  is decreased by the same amount $1/M$, 
thus keeping the correct normalization of the rows of the verbalization matrix. (The possibility that $h=h^*$ is not ruled out.) 
This procedure which is inspired by Moran's model of population genetics (Ewens, 2004) guarantees a minimum disturbance in the 
verbalization matrix and can be interpreted as the lateral inhibition of the competing word-object associations.  
We note that during the learning stage the agent playing the hearer role does not need to guess which object in the context is 
named by word $h^*$.

An extra rule is needed to keep the entries $p_{nh}^J$  within the unit interval $\left [ 0,1 \right ]$: we assume that once 
an entry reaches the values  $p_{nh}^J = 1$ or $p_{nh}^J = 0$    it becomes fixed, so
the extremes of the unit interval act as absorbing barriers for the stochastic dynamics
of the learning algorithm.

\subsection{Supervised learning}

The setting is identical to that described before except that now the hearer must guess which object in the context the speaker 
named by $h^*$  and then communicate its choice to the speaker (using some nonlinguistic means, such as pointing to the chosen object). 
In turn, the speaker must provide another nonlinguistic hint to indicate which object in the context it named by word $h^*$. 
Let us assume that the speaker associates word  $h^*$ to object $n_1$. If the hearer's guess happens to be the correct one, 
then both entries   $p_{n_1 h^*}^I$ and $p_{n_1 h^*}^J$  are incremented by the amount $1/M$. 
Furthermore, two words, say $h_s$ and $h_h$, are chosen randomly and the entries $p_{n_1 h_s}^I$ and $p_{n_1 h_h}^J$
are decreased by $1/M$ so the normalization of row  $n_1$ is preserved in both verbalization matrices. 
Suppose now the hearer's guess is wrong, say, object $n_2$  instead of $n_1$. Then both entries  $p_{n_1 h^*}^I$ and $p_{n_2 h^*}^J$
are decreased by the amount $1/M$  and, as before, two words $h_s$ and $h_h$ are chosen randomly and the entries 
$p_{n_1 h_s}^I$ and $p_{n_2 h_h}^J$ are increased by $1/M$. As in the unsupervised case, the extremes 
$p_{nh}^{I,J} = 1$ and $p_{nh}^{I,J} = 0$  are absorbing barriers.

The weak point of this learning scheme is the need for nonlinguistic hints to communicate the success or failure 
of the communication episode. This implies that, prior to learning, the agents are already capable to communicate (and understand) 
sophisticated meanings such as success and failure and behave (by updating their verbalization matrices) accordingly. 
In fact, feedback about the outcome of the communication episode may be seen as a form of telepathic meaning transfer.

\section{Results}

Simulation experiments of the two learning algorithms described above show, not surprisingly, that after a transient the 
two agents become identical, in the sense that they are described by the same verbalization matrix. In addition, in the case of 
unsupervised learning the stochastic dynamics always leads to binary verbalization matrices, i.e., matrices whose entries  $p_{nh}$
can take on the values 1 or 0 only. Of course, once the dynamics produces a binary matrix it becomes frozen. This same outcome 
characterizes the supervised case as well, except in the cases that the lexicon size $H$  is on the same order of the context size $C$. 
However, as we focus on the regime where  $C$ is finite and $N$ and $H$  are large we can guarantee that the stochastic dynamics 
leads to binary verbalization matrices regardless of the learning procedure.

Once the dynamics becomes frozen (and so the learning stage is over) we measure the average communication error $\epsilon$  as follows. 
The speaker chooses object $n$ from the list of  $N$ objects and emits the corresponding word 
(there is a unique word assigned to any given object, i.e., there is a single entry 1 in any row of the verbalization matrix). 
The hearer must then infer which object is named by that word. Since the same word can name many objects 
(i.e., there may be many entries 1 in a given column), the probability $\phi_n$  that the hearer's guess is correct is simply the 
reciprocal of the number of objects named by that word. This probability is the communication accuracy regarding object $n$. 
The procedure is repeated for the  $N$   objects, so  the average communication
error is defined as  $\epsilon = 1 - \phi$ where  $\phi = \sum_n \phi_n/N$  is the average communication accuracy of the algorithm.

As already pointed out, the normalization condition on the rows of the verbalization matrix  $P$ allows for the possibility 
that a certain number of words are not used by the lexicon acquisition algorithms. Let $H_u \leq H$ stand for the actual number of words used by 
those algorithms. Then we can easily convince ourselves that  $H_u = \sum_n^N \phi_n $ simply by noting that $\sum_n' \phi_n = 1$
when the sum is restricted to objects that are 
associated to the same word. Finally, we note that in the definitions of these communication measures the context plays no role at all; 
indeed the context is relevant only during the learning stage.

It is important to estimate the optimal (minimum) communication error $\epsilon_m$  in our learning scenario since, 
in addition to being a lower bound to the communication error produced by the learning algorithms, it allows us to rate their absolute 
performances. For  $H \leq N$ the optimal communication error is obtained by making a one-to-one assignment between $H-1$ words and $H-1$ 
objects, and then assigning the single remaining word to the remaining $N-H+1$  objects. 
This procedure yields $\epsilon_m = 1 - H/N = 1 - \alpha$. 
For $ H > N$ we can obtain $\epsilon_m =0$ simply by discarding $H-N$ words and making a one-to-one word-object assignment with the 
other $N$  words. In fact, using our finding that $\phi = H_u/N$  we see that, as expected, the optimal performance is obtained 
by setting $H_u = H$  if $H \leq N$  and  $H_u = N$ if $H > N$.

%------------------------------------------------------------------------
\begin{figure}
\centerline{\epsfig{width=0.52\textwidth,file=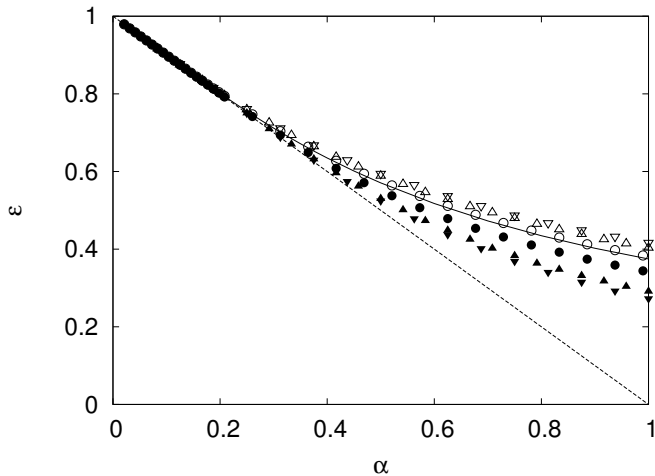}}
\par
\caption{Communication error $\epsilon$ as function of the ratio  $\alpha = H/N$ between the number of words $H$ and 
the number of objects $N$ for $N=16 (\bigtriangledown ), 24 (\triangle)$ and $96 (\bigcirc)$. The open (filled) symbols
represent the data for the unsupervised (supervised) algorithm. The error bars are smaller than the symbol sizes.
The solid line is the result of the
extrapolation for $N \to \infty$ (see Fig. \ref{fig:2})
whereas the dashed line represents the optimal performance
$1- \alpha$. The parameters are $C=2$ and $M=10^4$.
}
\label{fig:1}
\end{figure}
%------------------------------------------------------------------------------------

%------------------------------------------------------------------------
\begin{figure}
\centerline{\epsfig{width=0.52\textwidth,file=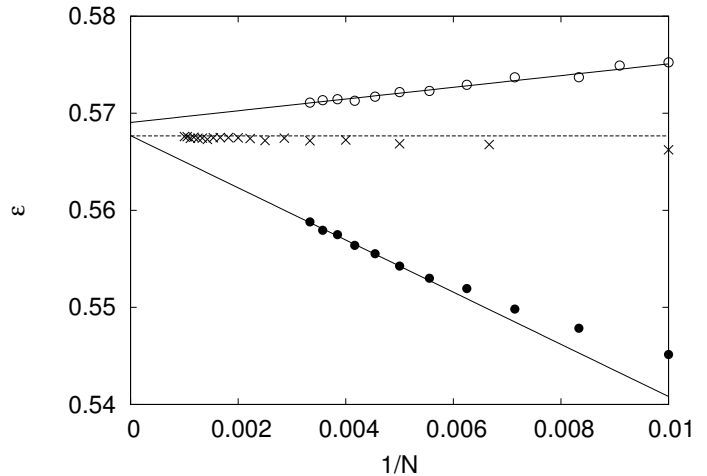}}
\par
\caption{Dependence of the communication error $\epsilon$ on the reciprocal of the
number of objects $1/N$ for $\alpha = 0.5$ for the unsupervised ($\circ$)
and supervised ($\bullet$) learning algorithms. The error bars are smaller than the symbol sizes.
The linear fittings (solid straight lines) yield
$ \epsilon = 0.5690 \pm 0.0003$ (unsupervised) and $\epsilon = 0.5677 \pm 0.0004$ (supervised) 
for $N \to \infty$. The Monte Carlo estimate of the error for the random assignment of objects to words 
is given by the symbols $\times$ and the dashed horizontal line corresponds to the
estimate of Eq.\ (\ref{Er}), $\epsilon_r = 0.5677$. The parameters are $C=2$ and $M=3~10^4$.
}
\label{fig:2}
\end{figure}
%------------------------------------------------------------------------

Figure \ref{fig:1} shows the comparison between the optimal performance and the actual performances of the two 
learning algorithms as function of the ratio $\alpha$. In this, as well as in the other figures of this paper, 
each symbol stands for the average over  $10^4$ independent samples or language games. The performance of the supervised 
algorithm deteriorates as the number of objects $N$ increases, in contrast to that of the unsupervised algorithm which actually shows a 
slight improvement in this case.  For $N \to \infty$, both algorithms produce the same communication error  
(see Fig.\ \ref{fig:2}), which is shown by the solid line in Fig.\ \ref{fig:1}.  We note that a preliminary comparative analysis 
of these algorithms for $N=8$  led to an incorrect claim about the general superiority of the supervised learning scheme 
(Fontanari \& Perlovsky, 2006). For small values of $\alpha$  the performances of the two learning algorithms are practically 
indistinguishable from the optimal performance, but as we will argue below the algorithms actually never achieve that performance, 
except for $\alpha=0$.

It is instructive to calculate the communication error in the case  that the  $N$ objects are 
assigned randomly to the $H$ words. This 
is a classical occupancy problem discussed at length in the celebrated book by Feller (1968).
In this occupancy problem, the probability $P_m$ that the number of words $m$  not used in the assignment of the $N$ objects to 
the $H$ words  (i.e., $m = H - H_u$) is 
\begin{equation}\label{m}
P_m = \left ( \begin{array}{c} H \\ m \end{array} \right ) 
\sum_{\nu =0}^{H-m} \left ( \begin{array}{c} H-m \\ \nu \end{array} \right ) \left ( -1 \right )^{\nu} 
\left ( 1 - \frac{m + \nu}{H} \right )^N ,
\end{equation}
which in the limits $N \to \infty $ and $H \to \infty$ reduces to the Poisson distribution 
\begin{equation}\label{poisson}
p \left (m; \lambda \right ) = \mbox{e}^{-\lambda} \frac{\lambda^m}{m!}
\end{equation}
where $\lambda = H \exp \left ( - N/H \right )$ remains bounded (Feller, 1968). Hence the average communication 
accuracy resulting from the random assignment of objects to words is simply 
$ \left ( H -  \left \langle m \right \rangle \right )/N$, which yields the  communication error
\begin{equation}\label{Er}
\epsilon_r = 1 - \alpha + \alpha \mbox{e}^{-1/\alpha} .
\end{equation}
Surprisingly, this equation describes perfectly the communication error of the two learning algorithms in the
limit $N \to \infty$ (solid line in Fig.\ \ref{fig:1}).  We note that the (small) 
discrepancy observed in Fig.\ \ref{fig:2} for the extrapolated data of the unsupervised algorithm and the analytical 
prediction can be reduced to zero by decreasing the learning rate $1/M$.
Equation (\ref{Er})  explains also why the  performances of the
algorithms are practically indistinguishable from the optimal performance for small $\alpha$, since the 
difference between them vanishes as $\exp \left ( -1/\alpha \right )$. In addition, Eq.\ (\ref{Er}) 
shows that in the limit of large $\alpha$, the communication error vanishes as  $1/\alpha$.

%------------------------------------------------------------------------
\begin{figure}
\centerline{\epsfig{width=0.52\textwidth,file=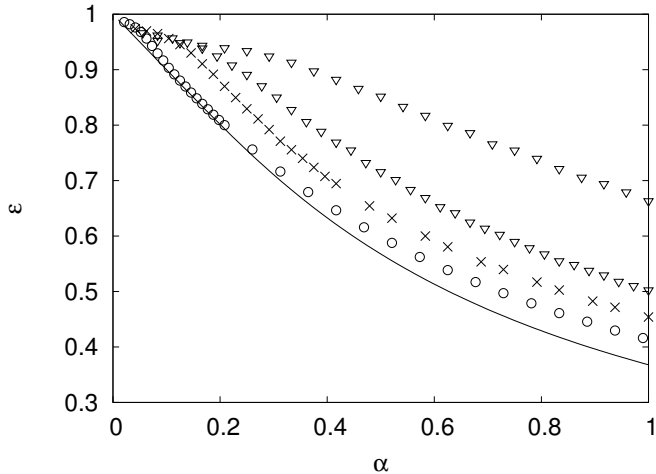}}
\par
\caption{Communication error $\epsilon$ of the unsupervised lexicon acquisition
algorithm for context size $C=4$ and $N=24 (\bigtriangledown ), 
36 (\triangle), 48 (\times)$, and $96 (\bigcirc)$. 
The error bars are smaller than the symbol sizes.
The learning rate is $1/M=10^{-4}$ and the solid line is the result of Eq.\ (\ref{Er}).
}
\label{fig:3}
\end{figure}
%------------------------------------------------------------------------

A word is in order about the effect of the context size $C$ on the performance of 
the two learning algorithms, since Figs.\ \ref{fig:1} and \ref{fig:2}
exhibit the results  for  $C=2$ only. Simulations for 
larger values of $C$ show that this parameter is completely irrelevant for the
performance of the supervised algorithm. Of course, this is expected since regardless
of the context size, at most two rows (object labels) of the verbalization matrices are updated. But
the situation is far from obvious for the unsupervised algorithm since $C$ determines
the number of rows to be updated in each round of the game. However, the results summarized
in Fig.\ \ref{fig:3} for $C=4$ indicate that, despite strong finite-size effects particularly
for small $\alpha$, the  communication error ultimately tends to $\epsilon_r$ in the
limit of large $N$. 
 
\section{Conclusion}

In this paper we have unveiled two remarkable results. First, the supervised and unsupervised schemes for 
bootstrapping a lexicon yield the same communication accuracy in the limit of very large lexicon sizes. 
For finite lexicon sizes the supervised scheme always outperforms the unsupervised one, but its performance degrades as the 
lexicon size increases, whereas the performance of the unsupervised learning algorithm improves slightly with increasing lexicon size 
(see Fig.\ \ref{fig:1}). Second, those performances tend to the communication accuracy obtained by a random occupancy problem in 
which the $N$  objects are assigned randomly to the $H$ words. These findings reveal a surprising inefficiency of traditional 
lexicon bootstrapping scenarios when evaluated in the realistic regime of very large lexicon sizes. 
It would be most interesting to devise sensible scenarios that reproduce the optimal communication performance or, at least, 
that exhibit an communication error that decays faster than the random occupancy result, $1/\alpha=N/H$, in the case the number of 
available words is much greater than the number of objects ($H \gg N$).

The scenarios studied here are easily adapted to model the problem of lexicon acquisition (rather than bootstrapping): 
we have just to assume that one of the agents, named the master in this case, knows the correct lexicon and so its verbalization 
matrix is kept fixed during the entire learning procedure; the verbalization matrix of the other agent -- the pupil -- is allowed to 
change following the update algorithms described before (see, e.g., Fontanari, Tikhanoff,  Cangelosi, Ilin, \& Perlovsky, 2009). 
Most interestingly, in this context, statistical world learning has been observed in controlled experiments involving infants 
(Smith \& Yu, 2008) and adults (Yu \& Smith, 2007). Similar experiments, but now aiming at bootstrapping a lexicon, could be 
easily carried out by replacing our virtual agents by two adults, who would then resort to some conscious or unconscious 
mechanism to track the co-occurrence of words and objects. Of course, the very emergence of pidgin - a means of communication 
between two or more groups which lack a common language (Thomason \& Kaufman, 1988) - can be seen as a realization of such 
an experiment and serves as additional justification for the study of lexicon bootstrapping.

\section*{Acknowledgments}

The research at S\~ao Carlos was supported in part by CNPq, FAPESP and SOARD grant FA9550-10-1-0006. J.F.F. thanks 
the hospitality of the Adaptive Behaviour \& Cognition Research Group, University of Plymouth, where this research was initiated. 
The visit was supported by euCognition.org travel grant NA-097-6. Cangelosi also acknowledges the contribution of the ITALK project 
from the European Commission (FP7 ICT Cognitive Systems and Robotics).

\section*{References}

Baronchelli, A., Felici, M., Loreto, V.,  Caglioli, E., \& Steels, L. (2006).
Sharp transition towards shared vocabularies in multi-agent systems. {\it Journal of 
Statistical Mechanics}, P06014.

Bates, E., \& Elman, J. (1996). Learning rediscovered. {\it Science}, 274, 1849-1850.

Bloom, P. (2000). {\it How children learn the meaning of words}. Cambridge, MA: MIT
Press.

Brighton, H., Smith, K., \& Kirby, S. (2005). Language as an evolutionary system.
{\it Physics of Life Reviews}, 2, 177-226.

De Beule, J., De Vylder, B., \& Belpaeme, T. (2006). A cross-situational learning
algorithm for damping homonymy in the guessing game. In L.M. Rocha, M. Bedau, D. Floreano, R. Goldstone, A. Vespignani, \& L. Yaeger (Eds.), 
{\it Proceedings of the Xth Conference on Artificial Life} (pp. 466-472). Cambridge, MA: MIT Press.

Cangelosi, A. (2001). Evolution of Communication and Language using Signals, 
Symbols and Words. {\it IEEE Transactions on Evolutionary Computation}, 5, 93-101.

Dawkins, R., \& Krebs, J.R. (1978). Animal signals: information or manipulation? In:
J.R. Krebs, \& N. B. Davies (Eds.), {\it Behavioural ecology: an evolutionary approach} (pp. 282-309). Oxford, UK: Blackwel Scientific Publications.

Donald, M. (1991). {\it Origins of the Modern Mind}. Cambridge, MA: Harvard University
Press.

Ewens, W.J. (2004). {\it Mathematical Population Genetics}. New York: Springer-Verlag. 

Feller, W. (1968). {\it An Introduction to Probability Theory and Its Applications}.  Vol. I,
3rd Edition. New York: Wiley.

Fontanari, J.F. (2006). Statistical analysis of discrimination games. {\it European Physical
Journal B}, 54, 127-130.

Fontanari, J.F., \&  Perlovsky, L.I. (2006). Meaning creation and communication in a
community of agents. In {\it Proceedings of the 2006 International Joint Conference on Neural Networks} 
(pp. 2892-2897). Piscataway, NJ: IEEE Press.

Fontanari, J.F., \& Perlovsky, L.I. (2007). Evolving compositionality in evolutionary 
language games. {\it IEEE Transactions on Evolutionary Computation}, 11, 758-769.

Fontanari, J.F., \& Perlovsky, L.I. (2008). A game theoretical approach to the evolution 
of structured communication codes. {\it Theory in Biosciences}, 127, 205-214.

Fontanari, J.F., Tikhanoff , V., Cangelosi, A., Ilin, R., \& Perlovsky, L.I. (2009). Cross-
situational learning of object-word mapping using Neural Modeling Fields. {\it Neural Networks}, 22, 579-585.

Gleitman, L. (1990). The structural sources of verb meanings. {\it Language Acquisition}, 1,
1-55.

Hurford, J.R. (1989). Biological evolution of the Saussurean sign as a component of the
language acquisition device. {\it Lingua}, 77, 187-222.

Ke, J, Minett, J.W., Au, C.-P., \& Wang, W.S.-Y. (2002). Self-organization and Selection
in the Emergence of Vocabulary. {\it Complexity}, 7, 41-54.

Kirby, S. (2002). Natural language from artificial life. {\it Artificial Life}, 8, 185-215.

Lenaerts, T., Jansen, B., Tuyls, K., \& De Vylder, B. (2005). The evolutionary language
game: An orthogonal approach. {\it Journal of Theoretical Biology}, 235, 566-582.

Nowak, M.A., \& Krakauer, D.C. (1999).The evolution of language. {\it Proceedings of the
National Academy of Sciences USA}, 96, 8028-8033.

Oliphant, M., \& Batali, J. (1997). Learning and the emergence of coordinated
communication, {\it Center for Research on Language Newsletter}, 11.

Pinker, S. (1984). {\it Language learnability and language development}. Cambridge, MA:
Harvard University Press.

Pinker, S., \& Bloom, P. (1990). Natural languages an natural selection, {\it Behavioral and 
Brain Sciences}, 13, 707-784.

Rosenthal, T., \& Zimmerman, B. (1978). {\it Social Learning and Cognition}. New York:
Academic Press.

Siskind, J.M. (1996). A computational study of cross-situational techniques for learning
word-to-meaning mappings. {\it Cognition}, 61, 39-91.

Smith, A.D.M. (2003a). Semantic generalization and the inference of meaning. 
{\it Lecture	Notes in Artificial Intelligence}, 2801, 499-506.

Smith, A.D.M. (2003b). Intelligent meaning creation in a clumpy world helps
communication. {\it Artificial Life}, 9, 557-574.

Smith, K., Kirby, S., \& Brighton, H. (2003). Iterated Learning: a framework for the
emergence of language. {\it Artificial Life}, 9, 371-386. 

Smith, K.,  Smith, A.D.M,  Blythe, R.A., \& Vogt, P. (2006). Cross-Situational 
Learning: A Mathematical Approach. {\it Lecture Notes in Computer Science}, 4211, 31-44.

Smith, L.B., \& Yu, C. (2008). Infants rapidly learn word-referent mappings via cross-
situational statistics. {\it Cognition}, 106, 1558-1568.

Steels, L. (1996). Perceptually grounded meaning creation. In M. Tokoro (Ed.),
{\it Proceedings of the Second International Conference on Multi-Agent Systems} (pp. 338-344). Menlo Park, CA: AAAI Press.

Steels, L., \& Kaplan, F. (1999). Situated grounded word semantics. In  {\it Proceedings of
the Sixteenth International Joint Conferences on Artificial Intelligence} (pp. 862-867). San Francisco, CA: Morgan Kauffman.

Steels, L. (2002). Grounding symbols through evolutionary language games. In A.
Cangelosi, \& D. Parisi (Eds.), {\it Simulating the Evolution of Language} (pp. 211-226). London: Springer-Verlag.

Steels, L. (2003). Evolving Grounded Communication for Robots. {\it Trends in Cognitive
Sciences}, 7, 308-312.

Thomason, S.G., \& Kaufman, T. (1988). {\it Language contact, creolization, and genetic 
linguistics}. Berkeley: University of California Press.

Yu, C., \& Smith, L.B (2007). Rapid word learning under uncertainty via cross-situational 
statistics. {\it Psychological Science}, 18, 414-420.

\end{document}